\documentclass[runningheads]{llncs}
\usepackage[T1]{fontenc}
\usepackage{graphicx}
\usepackage[misc]{ifsym} 
\usepackage{amssymb}
\usepackage{multirow}
\usepackage{booktabs}
\usepackage{color}
\usepackage[colorlinks,linkcolor=blue]{hyperref}
\newcommand{\repeatthanks}{\textsuperscript{\thefootnote}}

\begin{document}
\title{Deep Motion Network for Freehand 3D Ultrasound Reconstruction}

\author{
Mingyuan Luo\inst{1,2,3}\thanks{Mingyuan Luo and Xin Yang contribute equally to this work.} \and
Xin Yang\inst{1,2,3}\repeatthanks \and
Hongzhang Wang\inst{1,2,3} \and
Liwei Du\inst{1,2,3} \and
Dong Ni\inst{1,2,3}\textsuperscript{(\Letter)}
}


\authorrunning{M. Luo et al.}

\institute{
National-Regional Key Technology Engineering Laboratory for Medical Ultrasound, School of Biomedical Engineering, Health Science Center, Shenzhen University, China \\
\email{nidong@szu.edu.cn} \and
Medical Ultrasound Image Computing (MUSIC) Lab, Shenzhen University, China \and
Marshall Laboratory of Biomedical Engineering, Shenzhen University, China
}

\maketitle              

\begin{abstract}
Freehand 3D ultrasound (US) has important clinical value due to its low cost and unrestricted field of view. Recently deep learning algorithms have removed its dependence on bulky and expensive external positioning devices. However, improving reconstruction accuracy is still hampered by difficult elevational displacement estimation and large cumulative drift. In this context, we propose a novel deep motion network (MoNet) that integrates images and a lightweight sensor known as the inertial measurement unit (IMU) from a velocity perspective to alleviate the obstacles mentioned above. Our contribution is two-fold. First, we introduce IMU acceleration for the first time to estimate elevational displacements outside the plane. We propose a temporal and multi-branch structure to mine the valuable information of low signal-to-noise ratio (SNR) acceleration. Second, we propose a multi-modal online self-supervised strategy that leverages IMU information as weak labels for adaptive optimization to reduce drift errors and further ameliorate the impacts of acceleration noise. Experiments show that our proposed method achieves the superior reconstruction performance, exceeding state-of-the-art methods across the board.
\end{abstract}

\section{Introduction}
\begin{figure}[t]
\centering
\includegraphics[width=\textwidth]{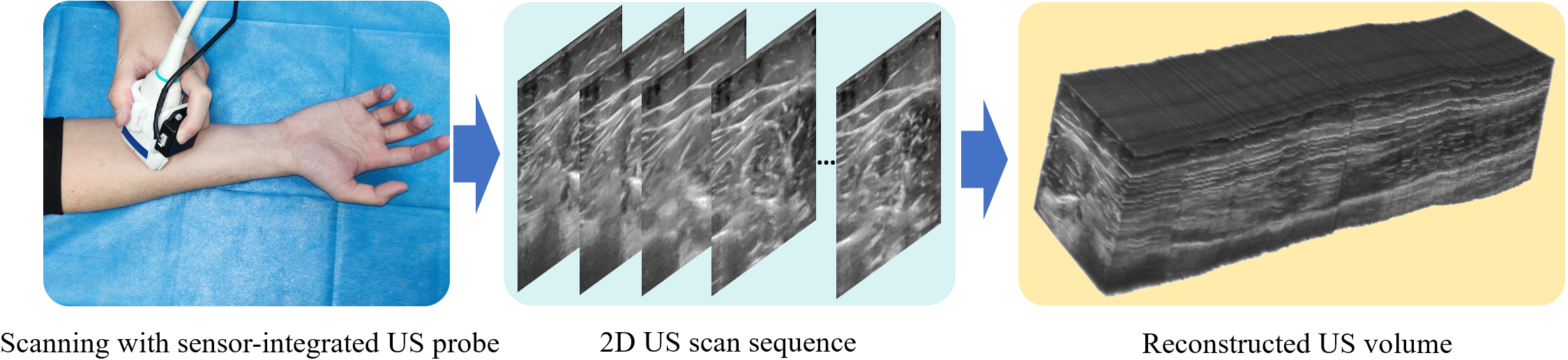}
\caption{Pipeline of freehand 3D US reconstruction with a lightweight inertial measurement unit (IMU) sensor.}
\label{fig:pipelile}
\end{figure}

Three-dimensional (3D) ultrasound (US) is widely used because of its intuitive visuals, easy interaction and rich clinical information. Freehand 3D US offers flexibility and simplicity over mechanical probes or electronic phased arrays~\cite{huang2017review}. It typically reconstructs the volume by calculating the relative positions of a series of US images. Recently, freehand reconstruction techniques have moved away from complex and costly external positioning systems previously used to obtain high-precision positions~\cite{prevost2017deep,prevost20183d,guo2020sensorless,luo2021self}. However, improving reconstruction accuracy remains difficult as these techniques only rely on images to infer relative positions. Specifically, the difficulty in estimating elevational displacement between images and the accumulation of drift errors make the reconstruction very challenging. In this regard, lightweight sensors are expected to avoid the disadvantages of traditional position sensors while improving the deep learning based reconstruction performance, as illustrated in Fig.~\ref{fig:pipelile}.

Freehand 3D US reconstruction has been studied for over half a century~\cite{mozaffari2017freehand}. Early solutions mainly relied on complex and expensive external positioning systems to accurately calculate image locations~\cite{mercier2005,mozaffari2017freehand,mohamed2019survey}. The non-sensor scheme is mainly speckle decorrelation~\cite{tuthill1998automated}, which uses speckle correlation between adjacent images to estimate relative motion and decomposes it into in-plane and out-of-plane parts. However, the reconstruction quality is susceptible to scan rate and angle~\cite{mohamed2019survey}. With the development of deep learning technology~\cite{lecun2015deep}, Prevost et al.~\cite{prevost2017deep} first used a convolutional neural network (CNN) to estimate the relative motion of US images. Guo et al.~\cite{guo2020sensorless} proposed a deep contextual learning network (DCL-Net) to mine the correlation information of US video clips for 3D reconstruction. Luo et al.~\cite{luo2021self} designed an online learning framework (OLF) that improves reconstruction performance by utilizing consistency constraints and shape priors. Although effective, the reconstruction performance based on deep learning still faces challenges such as difficult elevational displacement estimation and large cumulative drift.

Compared to complicated positioning systems, the lightweight sensor called inertial measurement unit (IMU) is inexpensive and takes up less space (Fig.~\ref{fig:pipelile}). IMU integrates triaxial accelerometer, gyroscope and magnetometer sensors to measure an object's triaxial orientation and acceleration~\cite{ahmad2013reviews}. Integrating an IMU does not increase US scanning complexity. Prevost et al.~\cite{prevost20183d} first demonstrated the promise of boosting reconstruction performance by simply concatenating the IMU orientation with the fully connected layer of neural networks, due to the low signal-to-noise ratio (SNR) of the IMU acceleration. 

In this study, we propose a novel lightweight sensor-based deep motion network (MoNet) for freehand 3D US reconstruction. Our contribution is two-fold. First, we equip a lightweight IMU sensor and exploit the IMU acceleration and orientation for the first time to improve image-based reconstruction performance. We propose a temporal and multi-branch structure from a velocity perspective to mine the valuable information of low-SNR acceleration. Second, we propose a multi-modal online self-supervised strategy that leverages IMU information as weak label for adaptive optimization to reduce drift errors and further amelioratethe impacts of acceleration noise. We thoroughly validate the efficacy and generalizability of MoNet on the collected arm and carotid scans. Experimental results show that MoNet achieves state-of-the-art reconstruction performance by fully utilizing the IMU information and deeply fusing it with image content.

\section{Methodology}

\begin{figure}[t]
\centering
\includegraphics[width=\textwidth]{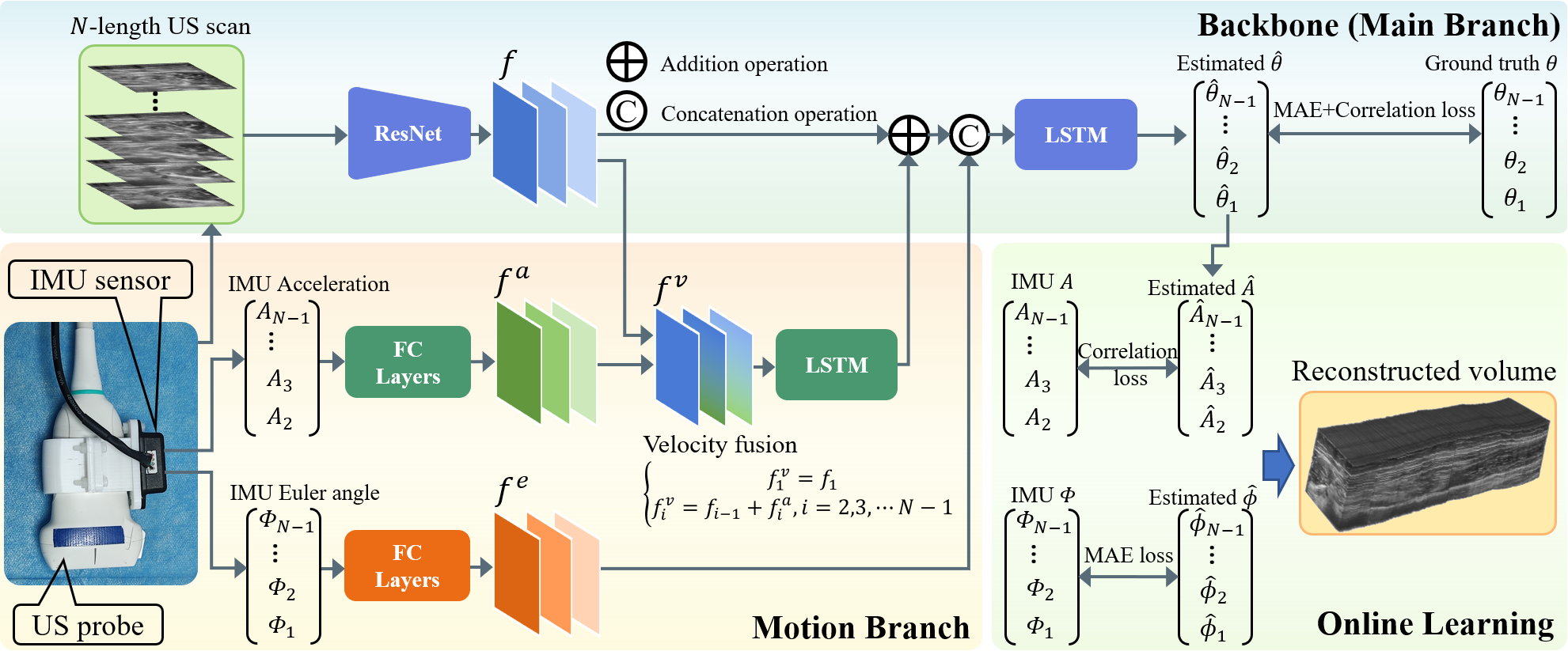}
\caption{Overview of our lightweight sensor-based deep motion network (MoNet).}
\label{fig:MoNet}
\end{figure}

We define the $N$-length scan as $I=\{I_i|i=1,2,\cdots,N\}$, and for image $I_i$, the corresponding orientation and acceleration vector provided by the IMU as $O_i=(O_x,O_y,O_z)_i$ and $A_i=(A_x,A_y,A_z)_i$, respectively. The $O_i$ is based on the east-north-up coordinate system, and the Euler angle $\Phi_i=(\Phi_x,\Phi_y,\Phi_z)_i$ between the images $I_i$ and $I_{i+1}$ can be calculated as 
\begin{equation}
\Phi_i=M^{-1}(M(O_i)^{-1}*M(O_{i+1})),\quad i=1,2,\cdots,N-1, 
\end{equation}
where $M(\cdot)$ converts the orientation vector into a $3\times3$ rotation matrix and $M^{-1}(\cdot)$ denotes the inverse operation of $M(\cdot)$. For the $A_i$, we first subtract the component of the gravity direction $g_i$ computed from $O_i$. We also adjust the mean value of acceleration to zero to reduce the influence of noise. The calculation can be expressed as 
\begin{equation}
A_i \gets (A_i-g_i)-\frac{1}{N}\sum_i{(A_i-g_i)},\quad i=1,2,\cdots,N.
\end{equation}

Fig.~\ref{fig:MoNet} illustrates the proposed lightweight sensor-based deep motion network (MoNet). MoNet takes all adjacent images $\{(I_i,I_{i+1})|i=1,2,\cdots,N-1\}$, the acceleration $\{A_i|i=2,3,\cdots,N-1\}$ and the Euler angle $\{\Phi_i|i=1,2,\cdots,N-1\}$ as inputs to estimate the relative transformation parameters $\theta=\{\theta_i|i=1,2,\cdots,N-1\}$, where $\theta_i$ indicates 3 translations $t_i=(t_x,t_y,t_z)_i$ and 3 rotation degrees $\phi_i=(\phi_x,\phi_y,\phi_z)_i$, respectively. It contains a temporal and multi-branch structure for mining valuable IMU information. Furthermore, the multi-modal online self-supervised strategy is used to improve estimation accuracy further.

\subsection{Temporal and Multi-branch Structure for IMU Fusion}
As shown in Fig.~\ref{fig:MoNet}, we construct the MoNet using ResNet~\cite{he2016deep} and LSTM~\cite{hochreiter1997long}. ResNet is a powerful feature extraction network that is commonly used to design networks for various tasks. LSTM is used to process temporal information, memorizing the knowledge of all historical images as contextual information to help estimate future parameters.

Estimating out-of-plane elevational displacements from images alone is a huge challenge. In this study, we introduce IMU acceleration for the first time to address this challenge. We first argue that the output features of the ResNet with adjacent images as input implicitly contain  relative distance information between images. Second, scanning speed can be represented by relative distance, assuming a constant sampling time between adjacent images. Hence, the IMU acceleration should be associated with the features containing velocity information, allowing complementary learning through a multi-branch structure implementation.

Within the multi-branch structure as shown in Fig.~\ref{fig:MoNet}, we map the processed acceleration to a high-dimensional space using multiple fully connected layers and reshape it into 2D features for easy combination with image features. The acceleration feature $f^a$ is then added to the implied velocity feature $f$ to construct the IMU velocity feature $f^v$, which can be expressed as
\begin{equation}
\left\{
\begin{array}{lr}
f^v_1=f_1, &  \\
f^v_i=f_{i-1}+f^a_i, & i=2,3,\cdots,N-1.
\end{array}
\right.
\end{equation}
Subsequently, to minimize the impact of acceleration noise on network performance in short sequence intervals, we feed $f^v$ into a LSTM to enhance the velocity features using the temporal context information. The LSTM output is merged into the main branch and added to the ResNet output. By fusing the IMU acceleration, the multi-branch structure provides a better estimation of elevational displacements. Furthermore, the Euler angle calculated from IMU orientation is fed into multiple fully connected layers and concatenated to the LSTM input in the main branch to enhance the Euler angle estimation in the relative transformation parameters.

In the training phase, the loss function of MoNet contains two items. The first item minimizes the mean absolute error (MAE) between the estimated transformation parameters $\hat{\theta}$ and ground truth $\theta$. The second item is the Pearson correlation loss from~\cite{guo2020sensorless}, which aids in learning the general trend of the scan.
\begin{equation}
L=\|\hat\theta-\theta\|_1+(1-\frac{\textbf{Cov}(\hat\theta,\theta)}{\sigma(\hat\theta)\sigma(\theta)}),
\end{equation}
where $\left\|\cdot\right\|_1$ calculates L1 normalization, $\textbf{Cov}$ denotes the covariance, and $\sigma$ indicates the standard deviation.

\subsection{Multi-modal Online Self-supervised Strategy}
Most previous 3D reconstruction methods~\cite{prevost2017deep,prevost20183d,guo2020sensorless} relied solely on offline network training and direct online inference. This strategy often fails to handle scans with a different distribution than the training data. Online self-supervised learning can address this challenge by leveraging adaptive optimization and valuable prior knowledge. In this study, we propose a multi-modal online self-supervised strategy that leverages IMU information as weak labels for adaptive optimization to reduce drift errors and further ameliorate the impacts of acceleration noise. Specifically, as shown in Fig.~\ref{fig:MoNet}, the IMU acceleration and Euler angle are used as weak labels for iterative optimization after estimating the transformation parameters using the trained MoNet. The estimated acceleration $\hat{A}$ at the centroid of each image is calculated from the estimated translations $\hat{t}$ and scaled to match the mean-zeroed IMU acceleration.
\begin{equation}
    \hat{A}_i=(\hat{t}^{-1}_{i-1}+\hat{t}_{i})-\frac{1}{N-2}\sum_i(\hat{t}^{-1}_{i-1}+\hat{t}_{i}),\quad i=2,3,\cdots,N-1,
\end{equation}
where $\hat{t}^{-1}_{i-1}$ denotes the translations in the inversion of $\hat{\theta}_{i-1}$. The optimization loss $L_{online}$ is split into two components: acceleration and Euler angle. We measure the difference between the estimated acceleration $\hat{A}$ and IMU acceleration $A$ with the Pearson correlation loss. This exploits the correct trend of IMU acceleration over a wide range of each scan to improve the MoNet estimation and ameliorate the impacts of noise. Meanwhile, the MAE loss constrains the difference between the estimated Euler angle $\hat{\phi}$ and IMU Euler angle $\Phi$.
\begin{equation}
    L_{online}=(1-\frac{\textbf{Cov}(\hat{A},A)}{\sigma(\hat{A})\sigma(A)})+\|\hat\phi-\Phi\|_1
\end{equation}

\section{Experiments}
\paragraph{\textbf{Materials and Implementation.}}

\begin{figure}[t]
\centering
\includegraphics[width=\textwidth]{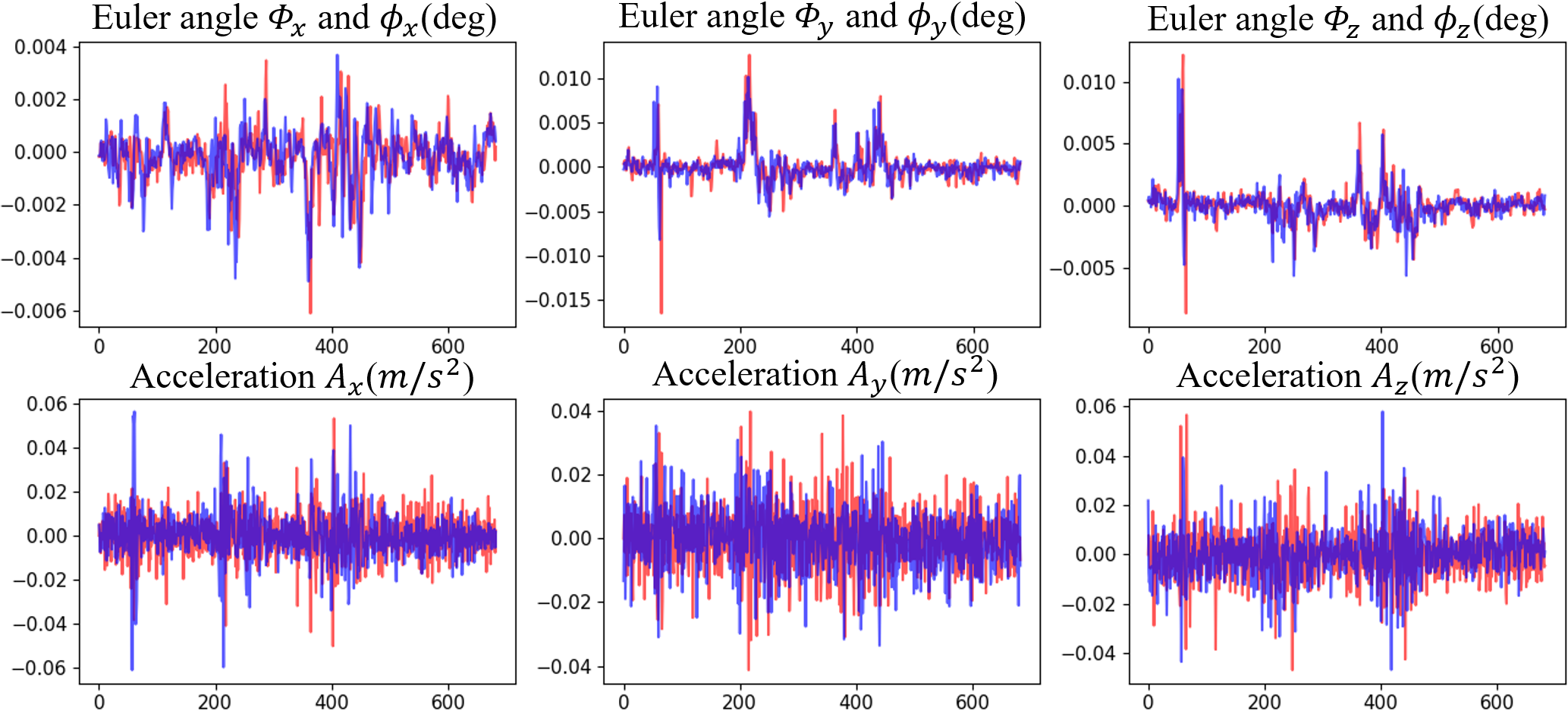}
\caption{Comparison of IMU data ($\Phi$ and $A$, blue line) and EM positioning data ($\phi$ and $A$, red line). The abscissa of each subfigure indicates the image index.}
\label{fig:noise}
\end{figure}

We built a data acquisition system to acquire all US scans and corresponding IMU data. The system consists of a portable US machine, an IMU sensor (WT901C-232, WitMotion ShenZhen Co.,Ltd., China) and an electromagnetic (EM) positioning transmitter/receiver. We acquired US images with a linear probe at 10 MHz and bound the IMU sensor to the probe to obtain the acceleration and orientation information. We also connected the EM receiver to the probe and utilized the EM positioning system to trace the scan route accurately. The US image depth is set as 3.5 cm, the IMU measurement resolution is $5\times10^{-4}$ g/LSB (acceleration) and 0.5 degree (orientation), and the EM positioning resolution is 1.4 mm (position) and 0.5 degree (orientation). We calibrated the entire system to acquire precise transformation parameters while minimizing EM positioning interference with the IMU. As shown in Fig.~\ref{fig:noise}, we analyzed the acceleration and Euler angle calculated by IMU data and compared them to the EM positioning. It can be observed that the acceleration is much more noisy at a single point compared to the Euler angle, but has a correct trend over a wide range.

We constructed two datasets including arm and carotid  using the acquisition system for evaluation. The arm dataset contains 250 scans from 41 volunteers with an average length of 94.83 mm. Scanning tactics included linear, curved, fast-and-slow and loop scans to mimic complex real-world situations. The carotid dataset contains 160 scans from 40 volunteers, with an average length of 53.71 mm. Only linear scans were performed due to the narrow pathway and the tissue's deformable tendency. The size of all scanned images is $478\times522$ pixel, and the image spacing is $0.075\times0.075$ mm$^2$. The local IRB approves the collection and use of data.

We randomly divided the arm and carotid datasets into 196/54 and 136/24 scans based on volunteer level to construct the training/test set. All images were scaled down to 0.6 times their original size. To increase the network's robustness and prevent overfitting, we conducted random augmentation of each scan, including subsequence intercepting, interval sampling, and sequence inversion. Each training scan was randomly augmented to 40 sequences and regenerated at each epoch to further improve the model generalizability. Each test scan was randomly augmented to 10 fixed sequences to mimic complex real-world situations. The Adam optimizer is used to optimize the MoNet. In the training phase, the epoch and batch size are 200 and 1, respectively. The learning rate is $10^{-4}$ and is cut in half every 30 epochs. During the online learning phase, the iteration epoch and learning rate are 60 and $2\times10^{-6}$, respectively. All code was implemented in PyTorch and ran on a RTX 3090 GPU.

\paragraph{\textbf{Quantitative and Qualitative Analysis.}}
To demonstrate the efficacy of our method, we performed the comparison with three SOTA approaches including CNN~\cite{prevost20183d}, DCL-Net~\cite{guo2020sensorless}, and OLF~\cite{luo2021self}. CNN used images, calculated optical flow and IMU Euler angles as input, whereas DCL-Net and OLF only used scanned images. Note that DCL-Net was modified to estimate the transformation parameters of adjacent images rather than the average parameters of several images to achieve better performance. Meanwhile, OLF's shape prior module was removed because muscles and vessels do not have prominent shape features. In this study, six criteria, including final drift rate (FDR), average drift rate (ADR), maximum drift (MD), sum of drift (SD), symmetric Hausdorff distance (HD), and mean error of angle (EA) are used to evaluate the performance (all criteria refer to~\cite{luo2021self} except EA). Furthermore, our ablation experiments compare Backbone (ResNet+LSTM), Backbone+IMU, and full MoNet (Backbone+IMU+Online) to validate the efficacy of IMU fusion and online learning strategy.

Table~\ref{tab:armcarotid} shows that our MoNet outperforms previous approaches on all metrics significantly ($t$-test, $p<0.05$) for both arm and carotid scans. We note that the CNN incorporating IMU Euler angles~\cite{prevost20183d} outperforms other sensorless approaches~\cite{guo2020sensorless,luo2021self} in terms of EA metric, verifying the effectiveness of IMU integration. However, the simple combination of CNN and IMU results in the lowest performance on other metrics. In addition, the OLF~\cite{luo2021self} and proposed MoNet outperform other methods~\cite{prevost20183d,guo2020sensorless} on most metrics, illustrating the importance of online learning. The ablation experiments further demonstrate that both IMU fusion and online learning strategy greatly improve the reconstruction accuracy.

\begin{table}[t]
\caption{The mean (std) results of different models on the arm and carotid scans. DCL: DCL-Net, Bk: Backbone. The best results are shown in blue.}
\begin{tabular}{l|c|c|c|c|c|c}
\toprule
\textbf{Models} & \textbf{FDR(\%)$\downarrow$} & \textbf{ADR(\%)$\downarrow$} & \textbf{MD(mm)$\downarrow$} & \textbf{SD(mm)$\downarrow$} & \textbf{HD(mm)$\downarrow$} & \textbf{EA(deg)$\downarrow$} \\ 
\hline
&\multicolumn{6}{c}{Arm scans} \\
\hline
CNN~\cite{prevost20183d} & 31.84(18.35)& 37.58(18.02)& 27.31(19.34)& 765.02(721.00)& 26.45(19.42)& 1.96(1.70) \\
DCL~\cite{guo2020sensorless} & 20.17(13.37)& 26.05(15.57)& 16.75(11.70)& 476.67(440.90)& 15.82(11.50)& 2.78(2.93) \\
OLF~\cite{luo2021self} & 15.00(13.09)& 20.91(12.23)& 12.69(8.30)& 374.07(350.84)& 11.92(8.02)& 2.38(2.25) \\
Bk & 16.42(14.24)& 22.40(16.13)& 13.10(10.15)& 394.90(427.16)& 11.92(9.22)& 2.29(2.50) \\
Bk+IMU & 14.05(10.36)& 20.12(12.63)& 11.56(7.54)& 352.54(334.08)& 10.76(7.20)& 1.75(1.57) \\
MoNet & \textcolor{blue}{12.75(9.05)} & \textcolor{blue}{19.05(11.46)} & \textcolor{blue}{10.24(7.36)} & \textcolor{blue}{332.29(316.36)} & \textcolor{blue}{9.40(7.13)} & \textcolor{blue}{1.55(1.46)} \\
\hline
&\multicolumn{6}{c}{Carotid scans} \\
\hline
CNN~\cite{prevost20183d} & 31.88(15.76)& 39.71(14.88)& 17.63(9.88)& 493.71(449.95)& 16.66(9.92)& 2.31(1.79) \\
DCL~\cite{guo2020sensorless} & 24.66(12.11)& 30.06(12.26)& 13.43(7.30)& 362.28(296.41)& 12.79(7.44)& 2.54(1.60) \\
OLF~\cite{luo2021self} & 20.08(14.72)& 29.21(14.99)& 11.01(6.76)& 326.91(317.36)& 10.38(6.53)& 2.59(1.58) \\
Bk & 20.55(18.73)& 30.74(19.09)& 11.06(7.84)& 318.38(306.10)& 10.03(6.73)& 2.61(1.72) \\
Bk+IMU & 17.78(11.50)& 27.47(13.05)& 9.73(4.83)& 285.99(239.24)& 9.14(4.83)& 2.18(1.43) \\
MoNet & \textcolor{blue}{15.67(8.37)}& \textcolor{blue}{25.08(9.34)}& \textcolor{blue}{8.89(4.31)}& \textcolor{blue}{258.83(208.12)}& \textcolor{blue}{8.28(4.29)}& \textcolor{blue}{1.50(0.98)} \\
\bottomrule
\end{tabular}
\label{tab:armcarotid}
\end{table}

\begin{figure}[t]
\centering
\includegraphics[width=\textwidth]{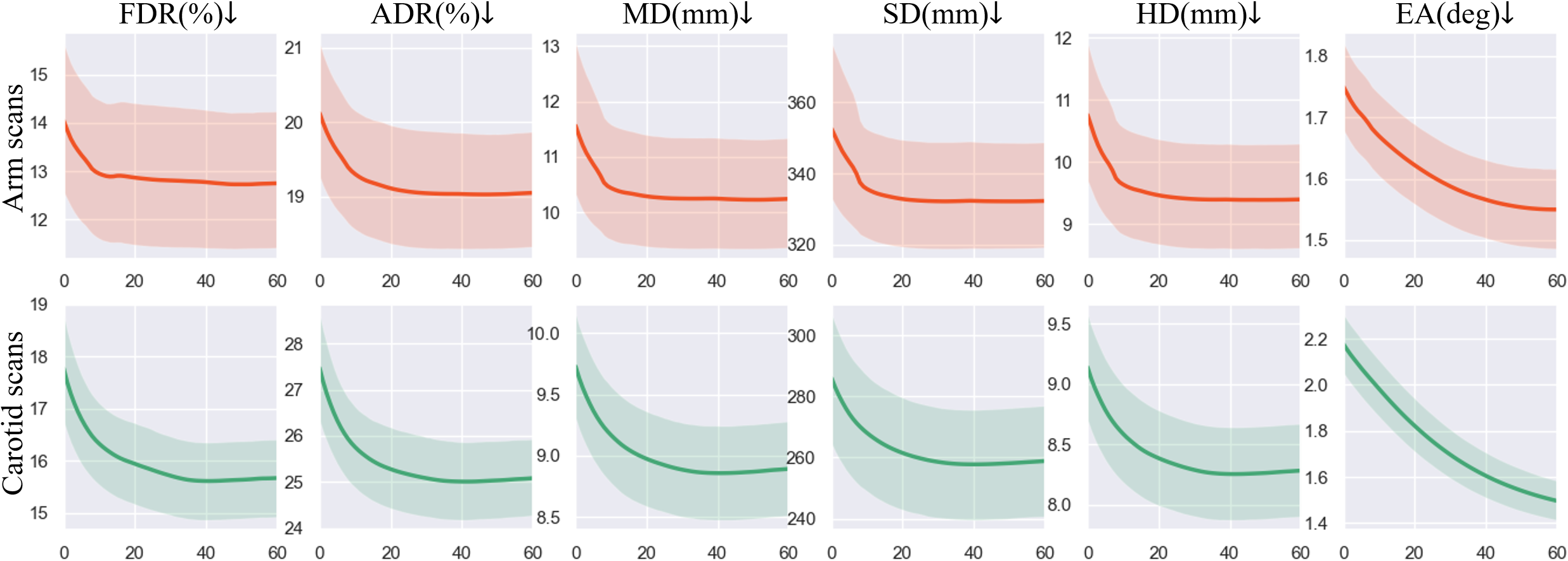}
\caption{Metric decline curves (with 95\% confidence interval) for online learning strategy. The abscissa and ordinate represent the number of iterations and the value of metrics.}
\label{fig:curve}
\end{figure}

\begin{figure}[t]
\centering
\includegraphics[width=\textwidth]{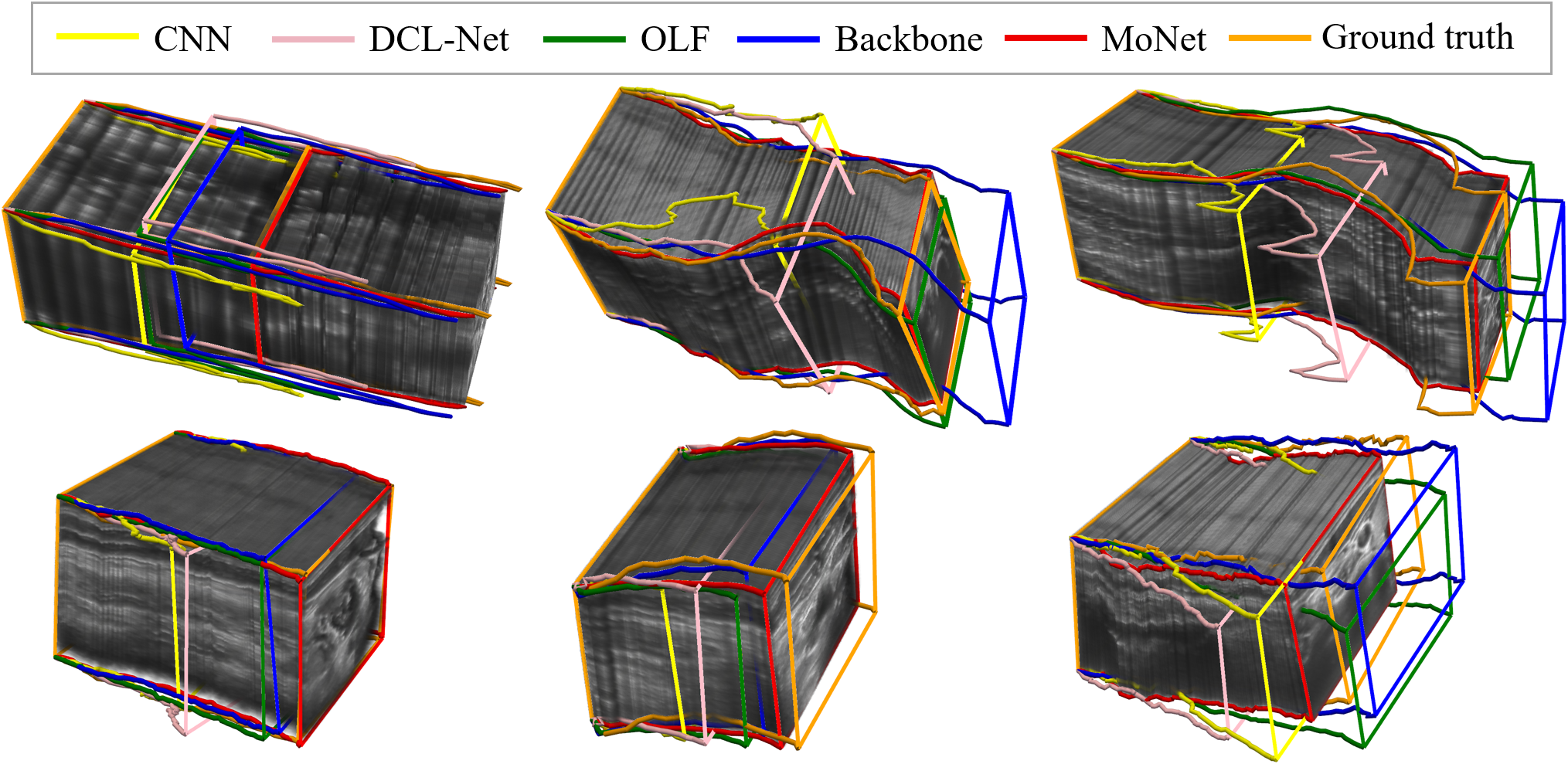}
\caption{Typical reconstruction cases on arm (Row I) and carotid (Row II) scans.}
\label{fig:reco}
\end{figure}

We further illustrate the metric decline curves when using online learning strategy in Fig.~\ref{fig:curve}. For both arm and carotid, the online learning substantially reduces all metrics and the most significant reductions are 12.64\% of HD (arm) and 31.19\% of EA (carotid). This further demonstrates the efficacy of our proposed online learning strategy. Fig.~\ref{fig:reco} visualizes six typical results to show the difference between our MoNet and other methods. It can be observed that our proposed MoNet achieves the closest reconstruction outcomes to ground truth compared to the Backbone or other methods.

\section{Conclusion}
We propose a lightweight sensor-based deep motion network (MoNet) to conduct freehand 3D US reconstruction. For the first time, we propose a temporal and multi-branch structure from a velocity perspective to mine the valuable information of IMU low-SNR acceleration. We propose a multi-modal online self-supervised strategy that leverages IMU information as weak labels for adaptive optimization to reduce drift errors and further ameliorate the impacts of acceleration noise. We build a data acquisition system and thoroughly validate the system efficacy on the arm and carotid datasets. Experiments show that MoNet achieves state-of-the-art reconstruction performance by fully utilizing the IMU information and deeply fusing it with image content. Future research will focus on extending this network to more anatomical structures.

\subsubsection*{Acknowledgements}
This work was supported by the grant from National Natural Science Foundation of China (Nos. 62171290, 62101343), Shenzhen-Hong Kong Joint Research Program (No. SGDX20201103095613036), and Shenzhen Science and Technology Innovations Committee (No. 20200812143441001).

%
%

\bibliographystyle{splncs04}
\bibliography{paper1226}

\end{document}